\documentclass[11pt,a4paper]{article}
\usepackage[hyperref]{emnlp2018}
\usepackage{times}
\usepackage{latexsym}
\usepackage{tabularx}
\usepackage{bigstrut}
\usepackage{multirow}
\usepackage{booktabs}
\usepackage{graphicx}

\usepackage{url}

\newcommand{\ignore}[1]{}

\title{On the Strength of Character Language Models for \\ Multilingual Named Entity Recognition} 

\author{Xiaodong Yu$^{\dag}$, Stephen Mayhew$^{*}$,  Mark Sammons$^{\dag}$, Dan Roth$^{*}$  \\
$^{\dag}$University of Illinois, Urbana-Champaign, \hspace{.2cm} $^{*}$University of Pennsylvania\\
  {\tt \small \{xyu71,mssammon\}@illinois.edu,  \{mayhew,danroth\}@seas.upenn.edu} 
}

\date{}

\aclfinalcopy

\begin{document}
\maketitle
\begin{abstract}
Character-level patterns have been widely used as features in English Named Entity Recognition (NER) systems. However, to date there has been no direct investigation of the inherent differences between name and non-name tokens in text, nor whether this property holds across multiple languages. 
This paper analyzes the capabilities of corpus-agnostic Character-level Language Models (CLMs) in the binary task of distinguishing name tokens from non-name tokens.  We demonstrate that CLMs provide a simple and powerful model for capturing these differences, identifying named entity tokens in a diverse set of languages at close to the performance of full NER systems. Moreover, by adding very simple CLM-based features we can significantly improve the performance of an off-the-shelf NER system for multiple languages.\footnote{The code and resources for this publication can be found at: \url{https://cogcomp.org/page/publication_view/846}}


\end{abstract}





\section{Introduction}

In English, there is strong empirical evidence that the character sequences that make up proper nouns tend to be distinctive.
Even divorced of context, a human reader can predict that  ``hoekstenberger'' is an entity, but ``abstractually''\footnote{Not a real name or a real word.} is not. Some NER research explores the use of character-level features including capitalization, prefixes and suffixes~\cite{CucerzanYa99,RatinovRo09}, and character-level models (CLMs)~\cite{KSNM03} to improve the performance of NER, but to date there has been no systematic study isolating the utility of CLMs in capturing distinctions between name and non-name tokens in English or across other languages.






\begin{figure}[t]
\centering
\includegraphics[scale=0.19]{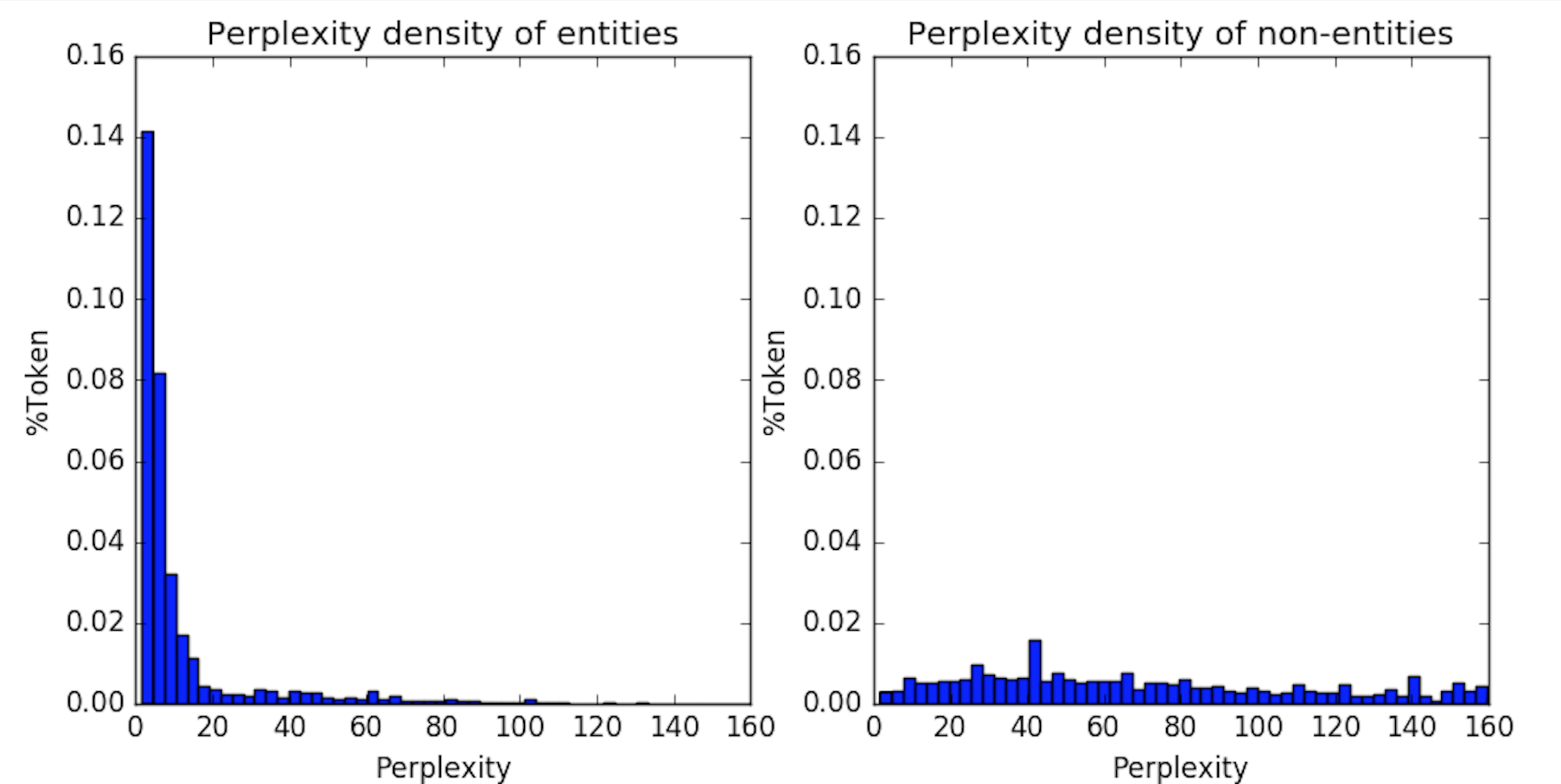} 
\caption{Perplexity histogram of entity (left) and non-entity tokens (right) in CoNLL Train calculated by entity CLM for both sides. The graphs show the percentage of tokens ($y$ axis) with different levels of CLM perplexities ($x$ axis). The entity CLM gives a low average perplexity and small variance to entity tokens (left), while giving non-entity tokens much higher perplexity and higher variance (right).} 

\label{ppxhist}
\end{figure}

We conduct an experimental assessment of the discriminative power of CLMs for a range of languages: English, Amharic, Arabic, Bengali, Farsi, Hindi, Somali, and Tagalog. These languages use a variety of scripts and orthographic conventions (for example, only three use capitalization), come from different language families, and vary in their morphological complexity. We demonstrate the effectiveness of CLMs in distinguishing name tokens from non-name tokens, as illustrated by Figure \ref{ppxhist}, which shows perplexity histograms from a CLM trained on entity tokens. Our models use only individual tokens, but perform extremely well in spite of taking no account of word context.

 
We then assess the utility of directly adding simple features based on this CLM implementation to an existing NER system, and show that they have a significant positive impact on performance across many of the languages we tried. By adding very simple CLM-based features to the system, our scores approach those of a state-of-the-art NER system \cite{LBSKD16} 
across multiple languages, demonstrating both the unique importance and the broad utility of this approach. 


\section{Methods}

\subsection{Character Language Models}
We propose a very simple model in which we train an entity CLM on a list of entity tokens, and a non-entity CLM on a list of non-entity tokens. Both lists are unordered, with all entries treated independently. Each token is split into characters and treated as a ``sentence'' where the characters are the ``words.''  For example, ``Obama" is an entity token, and is split into ``O b a m a".  From these examples we learn a score measuring how likely it is that a sequence of characters forms an entity.  At test time, we also split each word into characters and determine perplexity using the entity and non-entity CLMs. We assign the label corresponding to the lower perplexity CLM.

We experiment with four different kinds of language model: N-gram model, Skip-gram model, Continuous Bag-of-Words model (CBOW), and Log-Bilinear model (LB). We demonstrate that the N-gram model is best suited for this task. 

Following \newcite{PengRo16}, we implement N-gram using SRILM \cite{Stolcke02} with order 6 and Witten-Bell discounting.\footnote{We experimented with different orders on development data, but found little difference between them.} For Skip-Gram and CBOW CLMs, we use the Gensim implementation \cite{rehurek2010software} for training and inference, and we build the LB CLM using the OxLM toolkit \cite{baltescuBlHo2014}.

\begin{table}[t!]
\small
  \begin{center}
    \begin{tabular}{lrrrr}
    \toprule
    & \multicolumn{2}{c}{Train} & \multicolumn{2}{c}{Test} \\
    Language & Entity & Non-entity & Entity & Non-entity \\
    \midrule
    English & 29,450 & 170,524 & 7,194 & 38,554 \\
    Amharic & 5,886 & 46,641 & 2,077 & 16,235 \\
    Arabic & 7,640 & 52,968 & 1,754 & 15,073 \\
    Bengali & 15,288 & 108,592 & 4,573 & 32,929 \\
    Farsi & 4,547 & 50,084 & 1,608 & 13,968 \\
    Hindi & 5,565 & 69,267 & 1,947 & 23,853\\
    Somali & 6,467 & 51,034 & 1,967 & 14,545 \\
    Tagalog & 11,525 & 102,894 & 3,186 & 29,228 \\
    \bottomrule
    \end{tabular}
  \end{center}
  \caption{Data statistics for all languages, showing number of entity and non-entity tokens in Train and Test.}
\label{tab:datastats}
\end{table}

\subsection{Data}
\label{sec:data}

To determine whether name identifiability applies to languages other than English, 
we conduct experiments on a range of languages for which we had previously gathered resources (such as Brown clusters): English, Amharic, Arabic, Bengali, Farsi, Hindi, Somali, and Tagalog.

For English, we use the original splits from the ubiquitous CoNLL 2003 English dataset \cite{TjongKimSangMe03}, which is a newswire dataset annotated with Person (PER), Organization (ORG), Location (LOC) and Miscellaneous (MISC). To collect the list of entities and non-entities as the training data for the Entity and Non-Entity CLMs, we sample a large number of PER/ORG/LOC and non-entities from Wikipedia, using types derived from their corresponding FreeBase entities~\cite{LingWe12}. 


For all other languages, we use a subset of the corpora from the LORELEI project annotated for the NER task \cite{StrasselTr16}. We build our entity list using the tokens labeled as entities in the training data, and our non-entity list from the remaining tokens. These two lists are then used to train two CLMs, as described above.  


Our datasets vary in size of entity and non-entity tokens, as shown in Table \ref{tab:datastats}. The smallest, Farsi, has 4.5K entity and 50K non-entity tokens; the largest, English, has 29K entity and 170K non-entity tokens.

\begin{table*}[!ht]
\centering
\begin{tabular}{lr|rrrrrrrr}
\toprule
Model & \textbf{eng} & \textbf{amh}&\textbf{ara}&\textbf{ben}& \textbf{fas}& \textbf{hin} & \textbf{som} & \textbf{tgl} & avg \\
\midrule
Exact Match & 43.4 & 54.4 & 29.3 & 47.7 & 30.5 & 30.9 & 46.0 & 23.7& 37.5 \\
Capitalization & 79.5 &- & - & - & - & - & 69.5 & 77.6 & - \\
\midrule
SRILM & \textbf{92.8} & \textbf{69.9} & \textbf{54.7} & \textbf{79.4} & \textbf{60.8} & \textbf{63.8} & \textbf{84.1} & \textbf{80.5} & \textbf{70.5} \\
Skip-gram & 76.0 & 53.0 & 29.7 & 41.4 & 30.8 & 29.0 & 51.1 & 61.5 & 42.4  \\
CBOW & 73.7 & 50.0 & 28.1 & 40.6 & 32.6 & 26.5 & 56.4 & 62.5 & 42.4  \\
Log-Bilinear & 82.8 & 64.5 & 46.1 & 70.8 & 50.4 & 54.8 & 78.1 & 74.9 & 62.8\\
\midrule
CogCompNER (ceiling) & 96.5 & 73.8 & 64.9 & 80.6 & 64.1 & 75.9 & 89.4 & 88.6 & 76.8\\
\newcite{LBSKD16} (ceiling) & 96.4 & 84.4 & 69.8 & 87.6 & 76.4 & 86.3 & 90.9 & 91.2 & 83.8\\
\bottomrule
\end{tabular}
\caption{Token level identification F1 scores. Averages are computed over all languages other than English. Two baselines are also compared here: \textit{Capitalization} tags a token in test as entity if it is capitalized; and \textit{Exact Match} keeps track of entities seen in training, tagging tokens in Test that exactly match some entity in Train. The bottom section shows state-of-the-art models which use complex features for names, including contextual information. Languages in order are: English, Amharic, Arabic, Bengali, Farsi, Hindi, Somali, and Tagalog. The rightmost column is the average of all columns excluding English.}
\label{tab:neiAllLanguages}
\end{table*}

\section{CLM for Named Entity Identification}
\label{sec:clmForNei}
In this section, we first show the power of CLMs for distinguishing between entity and non-entity tokens in English, and then that this power is robust across a variety of languages. 

We refer to this task as Named Entity Identification (NEI), because we are concerned only with finding an entity span, not its label. We differentiate it from Named Entity Recognition (NER), in which both span and label are required. To avoid complicating this straightforward approach by requiring a separate mention detection step, we evaluate at the token-level, as opposed to the more common phrase-level evaluation. We also apply one heuristic: if a word has length 1, we automatically predict `O' (or non-entity). This captures most punctuation and words like `I' and `a'.


Figure 1 shows that for the majority of entity tokens, the entity CLM computes a relatively low perplexity compared to non-entity tokens. Though there also exist some non-entities with low entity CLM perplexity, we can still reliably identify a large proportion of non-entity words by setting a threshold value for entity CLM perplexity. If a token perplexity lies above this threshold, we label it as a non-entity token. The threshold is tuned on development data.

Since we also build a CLM for non-entities, we can also compare the entity and non-entity perplexity scores for a token. For those tokens not excluded using the threshold as described above, we compare the perplexity scores of the two models and assign the label corresponding to the model yielding the lower score.

We compare SRILM against Skip-gram and CBOW, as implemented in Gensim, and the Log-Bilinear (LB) model. We trained both CBOW and Skip-gram with window size 3, and size 20. We tuned LB, and report results with embedding size 150, and learning rate 0.1. Despite tuning the neural models, the simple N-gram model outperforms them significantly, perhaps because of the relatively small amount of training data.\footnote{We also tried a simple RNN+GRU language model, but found that the results were underwhelming.}

We compare the CLM's Entity Identification against two state-of-the-art NER systems: CogCompNER \cite{2018_lrec_cogcompnlp} and LSTM-CRF \cite{LBSKD16}. We train the NER systems as usual, but at test time we convert all predictions into binary token-level annotations to get the final score. As Table \ref{tab:neiAllLanguages} shows, the result of N-gram CLM, which yields the highest performance, is remarkably close to the result of state-of-the-art NER systems (especially for English) given the simplicity of the model.



\begin{table*}[ht]
\centering
\begin{tabular}{llr|rrrrrrrr}
\toprule
Model &  & \textbf{eng} & \textbf{amh}&\textbf{ara}&\textbf{ben}& \textbf{fas}& \textbf{hin} & \textbf{som} & \textbf{tgl} & avg \\
\midrule
\multirow{2}{*}{\newcite{LBSKD16}} & Full & 90.94 & \textbf{73.2} & 57.2 & \textbf{77.7} & \textbf{61.2} & \textbf{77.7} & 81.3 & \textbf{83.2} & \textbf{73.1} \\ 
& Unseen &  86.11& 51.9 & 30.2 & 57.9 & 41.4 & 62.2 & 66.5 & 72.8 & 54.7\\ 
\midrule
\multirow{2}{*}{CogCompNER} & Full & 90.88 &  67.5 & 54.8 & 74.5 & 57.8 & 73.5 & 82.0 & 80.9 & 70.1\\
& Unseen &  84.40 & 42.7 & 25.0 & 51.9 & 31.5 & 53.9 & 67.2 & 68.3 & 48.6 \\
\multirow{2}{*}{CogCompNER+LM} & Full & \textbf{91.21} & 71.3 & 
\textbf{59.1} & 75.5 & 59.0 & 74.2 & \textbf{82.1} & 78.5 & 71.4 \\
& Unseen & 85.20 & 48.4 & 32.0 & 54.0 & 31.2 & 55.4 & 68.0 & 65.2 & 50.6 \\
\bottomrule
\end{tabular}
\caption{NER results on 8 languages show that even a simplistic addition of CLM features to a standard NER model boosts performance. CogCompNER is run with standard features, including Brown clusters; \cite{LBSKD16} is run with default parameters and pre-trained embeddings. \textit{Unseen} refers to performance on named entities in Test that were not seen in the training data. \textit{Full} is performance on all entities in Test. Averages are computed over all languages other than English.} 
\label{tab:nerPlusNeiAllLanguages}
\end{table*}



\section{Improving NER with CLM features}
In this section we show that we can augment a standard NER system with simple features based on our entity/non-entity CLMs to improve performance in many languages. Based on their superior performance as reported in Section~\ref{sec:clmForNei}, we use the N-gram CLMs.  

\subsection{Features}
We define three simple features that capture information provided by CLMs and which we expect to be useful for NER.

\vspace*{0.1cm}
\textbf{Entity Feature~~} We define one ``isEntity" feature based on the perplexities of the entity and non-entity CLMs. We compare the perplexity calculated by entity CLM and non-entity CLM described in Section~\ref{sec:clmForNei}, and return a boolean value indicating whether the entity CLM score is lower. 



\vspace*{0.1cm}
\textbf{Language Features~~} We define two language-related features: ``isArabic" and ``isRussian". We observe that there are many names in English text that originate from other languages, resulting in very different orthography than native English names. We therefore build two language-based CLMs for Arabic and Russian. We collect a list of Arabic names and a list of Russian names by scraping name-related websites, and train an Arabic CLM and a Russian CLM. For each token, when the perplexity of either the Arabic or the Russian CLM is lower than the perplexity of the Non-Entity CLM, we return True, indicating that this entity is likely to be a name from Arabic/Russian. Otherwise, we return False.

\subsection{Experiments}
\label{sec:experiments}
We use CogCompNER \cite{2018_lrec_cogcompnlp} as our baseline NER system because it allows easy integration of new features, and evaluate on the same datasets as before.
For English, we add all features described above. For other languages, due to the limited training data, we only use the ``isEntity'' feature.  We compare with the state-of-the-art character-level neural NER system of \cite{LBSKD16}, which inherently encodes comparable information to CLMs, as a way to investigate how much of that system's performance can be attributed directly to name-internal structure.

The results in Table~\ref{tab:nerPlusNeiAllLanguages} show that for six of the eight languages we studied, the baseline NER can be significantly improved by adding simple CLM features; for English and Arabic, it performs better even than the neural NER model of~\cite{LBSKD16}. For Tagalog, however, adding CLM features actually impairs system performance.  

In the same table, the rows marked ``unseen'' report systems' performance on named entities in Test that were not seen in the training data. This setting more directly assesses the robustness of a system to identify named entities in new data. By this measure, Farsi NER is not improved by name-only CLM features and Tagalog is impaired. Benefits for English, Hindi, and Somali are limited, but are quite significant for Amharic, Arabic, and Bengali. 



\section{Discussion}

Our results demonstrate the power of CLMs for recognizing named entity tokens in a diverse range of languages, and that in many cases they can improve off-the-shelf NER system performance even when integrated in a simplistic way. 

However, the results from Section~\ref{sec:experiments} show that this is not true for all languages, especially when only considering unseen entities in Test: Tagalog and Farsi do not  follow the trend for the other languages we assessed 
even though CLM performs well for Named Entity Identification. 



While the end-to-end model developed by \cite{LBSKD16} clearly includes information comparable to that in the CLM, it requires a fully annotated NER corpus, takes significant time and computational resources to train, and is non-trivial to integrate into a new NER system. The CLM approach captures a very large fraction of the entity/non-entity distinction capacity of full NER systems, and can be rapidly trained using only entity and non-entity token lists -- i.e., it is corpus-agnostic.  For some languages it can be used directly to improve NER performance; for others (such as Tagalog), the strong NEI performance indicates that while it does not immediately boost performance, it can ultimately be used to improve NER there too.  

\section{Related Work}

  
  
\newcite{CucerzanYa99} is one of the earliest works to use character-based features (character tries) for NER.
The approach of \newcite{KSNM03} was one of the original papers in the CoNLL 2003 NER shared task. Their approach, which ranked in the top 3 for both English and German shared tasks, used character-based features for NER. They do two experiments: one with a character-based HMM, another with using character n-grams as features to a maximum entropy model. The focus on character-level patterns is similar to our work, but without the specific exploration of language models alone.

Using character-based models similar to ours, \newcite{SmarrMa02} show that unseen noun phrases can be accurately classified into a small number of categories using only a character-based model independent of context. We tackle a somewhat more challenging task of distinguishing entities from non-entities. 
\newcite{LBSKD16} use character embeddings in an LSTM-CRF model. Their ablation studies show that character-level features improve performance significantly. 
 
We are not aware of any work that directly evaluates CLMs for identifying name tokens, nor of work that demonstrates the utility of character-level information for identifying names in multiple languages.

\section{Conclusions and Future Work}
We have shown, in a series of simple experiments, that in many languages names are identifiable by character patterns alone, and that character level patterns have strong potential for building better NER systems. 


In the future, we plan to make a more thorough analysis of reasons for the high variance in NER performance. In particular, we will study why it is possible, as with Tagalog, to have high Named Entity Identification results but lose points in NER.

\section*{Acknowledgements}
This work was supported by a grant from Google, and by Contract HR0011-15-2-0025 with the US Defense Advanced Research Projects Agency (DARPA). Approved for Public Release, Distribution Unlimited. The views expressed are those of the authors and do not reflect the official policy or position of the Department of Defense or the U.S. Government. We appreciate the helpful discussions and suggestions from Haoruo Peng and Qiang Ning, and from the anonymous EMNLP reviewers.

\bibliographystyle{acl_natbib}
\bibliography{acl2018,ccg,cited-long,new}

\begin{thebibliography}{}
\expandafter\ifx\csname natexlab\endcsname\relax\def\natexlab#1{#1}\fi

\bibitem[{Baltescu et~al.(2014)Baltescu, Blunsom, and Hoang}]{baltescuBlHo2014}
Paul Baltescu, Phil Blunsom, and Hieu Hoang. 2014.
\newblock
  \href{https://ufal.mff.cuni.cz/pbml/102/art-baltescu-blunsom-hoang.pdf}{Oxlm:
  A neural language modelling framework for machine translation}.
\newblock {\em The Prague Bulletin of Mathematical Linguistics\/}
  102(1):81--92.
\newblock
  \href{https://ufal.mff.cuni.cz/pbml/102/art-baltescu-blunsom-hoang.pdf}{https://ufal.mff.cuni.cz/pbml/102/art-baltescu-blunsom-hoang.pdf}.

\bibitem[{Cucerzan and Yarowsky(1999)}]{CucerzanYa99}
Silviu Cucerzan and David Yarowsky. 1999.
\newblock Language independent named entity recognition combining morphological
  and contextual evidence.
\newblock In {\em EMNLP\/}.

\bibitem[{Khashabi et~al.(2018)Khashabi, Sammons, Zhou, Redman,
  Christodoulopoulos, Srikumar, Rizzolo, Ratinov, Luo, Do, Tsai, Roy, Mayhew,
  Feng, Wieting, Yu, Song, Gupta, Upadhyay, Arivazhagan, Ning, Ling, and
  Roth}]{2018_lrec_cogcompnlp}
Daniel Khashabi, Mark Sammons, Ben Zhou, Tom Redman, Christos
  Christodoulopoulos, Vivek Srikumar, Nicholas Rizzolo, Lev Ratinov, Guanheng
  Luo, Quang Do, Chen-Tse Tsai, Subhro Roy, Stephen Mayhew, Zhili Feng, John
  Wieting, Xiaodong Yu, Yangqiu Song, Shashank Gupta, Shyam Upadhyay, Naveen
  Arivazhagan, Qiang Ning, Shaoshi Ling, and Dan Roth. 2018.
\newblock {CogCompNLP}: Your swiss army knife for nlp.
\newblock In {\em 11th Language Resources and Evaluation Conference\/}.

\bibitem[{Klein et~al.(2003)Klein, Smarr, Nguyen, and Manning}]{KSNM03}
Dan Klein, Joseph Smarr, Huy Nguyen, and Christopher~D. Manning. 2003.
\newblock Named entity recognition with character-level models.
\newblock In {\em CoNLL\/}.

\bibitem[{Lample et~al.(2016)Lample, Ballesteros, Subramanian, Kawakami, and
  Dyer}]{LBSKD16}
Guillaume Lample, Miguel Ballesteros, Sandeep~K Subramanian, Kazuya Kawakami,
  and Chris Dyer. 2016.
\newblock Neural architectures for named entity recognition.
\newblock In {\em HLT-NAACL\/}.

\bibitem[{Ling and Weld(2012)}]{LingWe12}
Xiao Ling and Daniel~S Weld. 2012.
\newblock
  \href{http://aiweb.cs.washington.edu/ai/pubs/ling-aaai12.pdf}{Fine-grained
  entity recognition}.
\newblock In {\em Proceedings of the National Conference on Artificial
  Intelligence (AAAI)\/}.
\newblock
  \href{http://aiweb.cs.washington.edu/ai/pubs/ling-aaai12.pdf}{http://aiweb.cs.washington.edu/ai/pubs/ling-aaai12.pdf}.

\bibitem[{Peng and Roth(2016)}]{PengRo16}
Haoruo Peng and Dan Roth. 2016.
\newblock \href{http://cogcomp.org/papers/PengRo16.pdf}{Two discourse driven
  language models for semantics}.
\newblock In {\em Proc. of the Annual Meeting of the Association for
  Computational Linguistics (ACL)\/}.
\newblock
  \href{http://cogcomp.org/papers/PengRo16.pdf}{http://cogcomp.org/papers/PengRo16.pdf}.

\bibitem[{Ratinov and Roth(2009)}]{RatinovRo09}
L.~Ratinov and D.~Roth. 2009.
\newblock \href{http://cogcomp.org/papers/RatinovRo09.pdf}{Design challenges
  and misconceptions in named entity recognition}.
\newblock In {\em Proc. of the Conference on Computational Natural Language
  Learning (CoNLL)\/}.
\newblock
  \href{http://cogcomp.org/papers/RatinovRo09.pdf}{http://cogcomp.org/papers/RatinovRo09.pdf}.

\bibitem[{Rehurek and Sojka(2010)}]{rehurek2010software}
Radim Rehurek and Petr Sojka. 2010.
\newblock Software framework for topic modelling with large corpora.
\newblock In {\em In Proceedings of the LREC 2010 Workshop on New Challenges
  for NLP Frameworks\/}. Citeseer.

\bibitem[{Sang and Meulder(2003)}]{TjongKimSangMe03}
Erik F. Tjong~Kim Sang and Fien~De Meulder. 2003.
\newblock Introduction to the conll-2003 shared task: Language-independent
  named entity recognition.
\newblock In {\em CoNLL\/}.

\bibitem[{Smarr and Manning(2002)}]{SmarrMa02}
Joseph Smarr and Christopher~D. Manning. 2002.
\newblock Classifying unknown proper noun phrases without context.

\bibitem[{Stolcke(2002)}]{Stolcke02}
Andreas Stolcke. 2002.
\newblock Srilm-an extensible language modeling toolkit.
\newblock In {\em Seventh international conference on spoken language
  processing\/}.

\bibitem[{Strassel and Tracey(2016)}]{StrasselTr16}
Stephanie Strassel and Jennifer Tracey. 2016.
\newblock Lorelei language packs: Data, tools, and resources for technology
  development in low resource languages.

\end{thebibliography}

\end{document}